\documentclass{article}

\PassOptionsToPackage{numbers, compress}{natbib}
    \usepackage[preprint]{main}

\usepackage[utf8]{inputenc} 
\usepackage[T1]{fontenc}    
\usepackage[colorlinks=true,breaklinks=true,bookmarks=false,citecolor=green]{hyperref}       
\usepackage{url}            
\usepackage{booktabs}       
\usepackage{amsfonts}       
\usepackage{nicefrac}       
\usepackage{microtype}      
\usepackage{xcolor}         
\usepackage{graphicx}
\usepackage{amsmath}
\usepackage{amssymb}
\usepackage{subcaption}
\usepackage{wrapfig}
\usepackage{pifont}
\usepackage{algorithmic}
\usepackage[ruled]{algorithm2e}
\usepackage{multirow}
\usepackage{fontawesome}
\usepackage{cleveref}    
\usepackage{tikz}
\usepackage{xspace}
\usepackage{multirow} 
\newcommand{\boldstartspace}[1]{\medskip\noindent\textbf{#1}}

\title{
AnimaSpark: A Feed-Forward Method for Animating Arbitrary 3D Objects
}

\author{
    Yiming Zhao\textsuperscript{1}, Haoyu Sun\textsuperscript{1},
    Aoyu Wang$^{\dag}$\textsuperscript{1}, \\
    \textsuperscript{1}{Bytedance} \quad
    \vspace{-.20in}
}

\begin{document}

\makeatletter
\let\@oldmaketitle\@maketitle
\renewcommand{\@maketitle}{\@oldmaketitle
 \centering
    \includegraphics[width=0.95\linewidth]{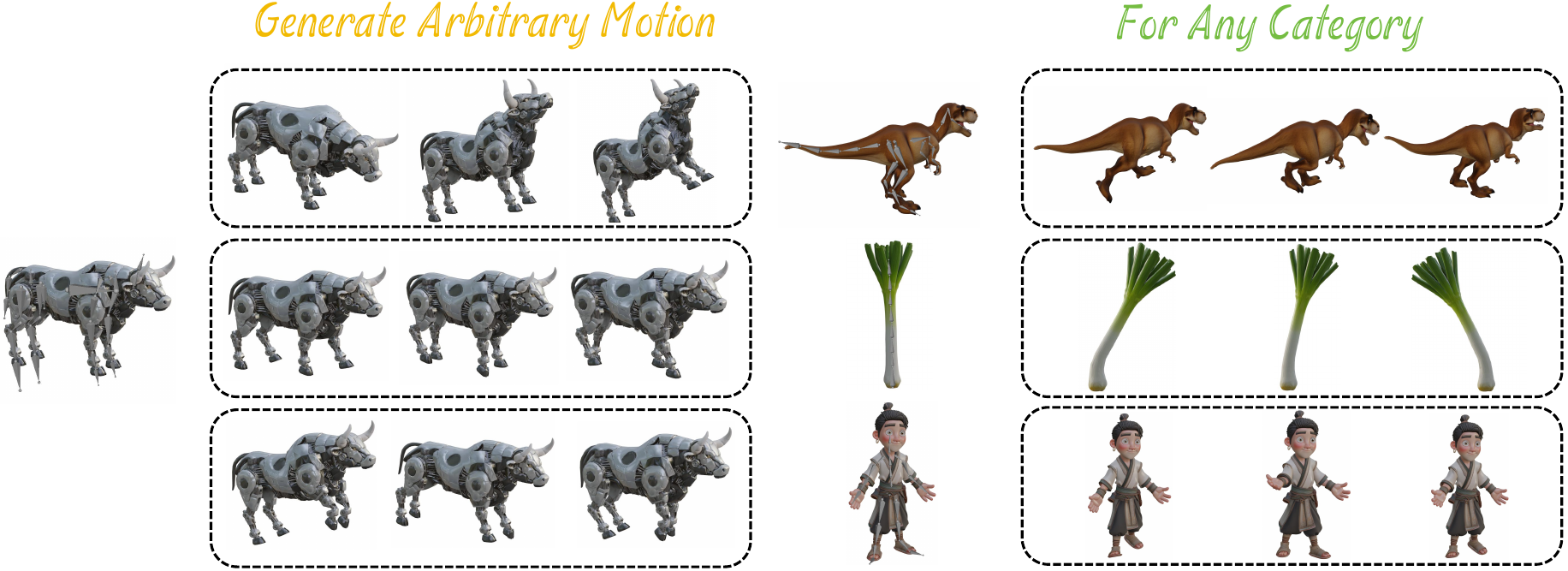}
  \captionof{figure}{Our method is capable of generating skeletal animations with diverse motions for 3D objects across a wide range of categories, including humanoids, animals, plants, and inanimate objects.
    }
  \label{teaser}
  \bigskip}
\makeatother

\maketitle

\let\oldthefootnote\thefootnote
\let\thefootnote\relax
\footnotetext{$^{\dag}$ Corresponding authors. Email: wangaoyu.0302@bytedance.com }
\let\thefootnote\oldthefootnote

\begin{abstract}
While recent advancements in generative AI have substantially accelerated static 3D model creation workflows, the synthesis of category-agnostic 3D animations remains a significant bottleneck in 3D asset production. Current methods for category-agnostic animation generation exhibit critical limitations in inference speed, motion quality, and adherence to textual prompts, thereby leaving the process dependent on labor-intensive manual artistry. To address these challenges, this paper introduces AnimaSpark, a novel pipeline for category-agnostic 3D animation generation. Our approach is motivated by the key insight that for many fundamental motions in the 3D world, the corresponding joint transformations can often be effectively modeled within a two-dimensional subspace. The pipeline begins by rendering a rigged static 3D model into multi-layered image representations of its mesh and skeleton, which are subsequently fed into a video generation model. We then employ a keypoint tracking algorithm on the generated video to capture the motion of the skeletal joints projected onto the camera's viewing plane. In the final stage, we distill the planar translations and rotations from these tracked keypoints and lift them from the 2D domain into 3D space to animate the character. Comprehensive evaluations reveal that our method achieves superior performance over existing state-of-the-art techniques across key metrics, including text-motion alignment, quality of motion, and computational efficiency.
\end{abstract}

\section{Introduction}
\label{sec:intro}

3D assets constitute the foundational building blocks for a myriad of cutting-edge industries, including game development, virtual and augmented reality (VR/AR) experiences, and embodied AI simulations. A complete production pipeline for these assets typically encompasses three sequential stages: the creation of a static 3D model, the rigging of the model with an internal skeleton, and finally, the generation of 3D animation. In recent years, generative AI-driven methods for static 3D model creation~\cite{tripo3d,xiang2025structured,lai2025hunyuan3d} have gained widespread industry acceptance and seen extensive practical adoption. Concurrently, template-free rigging methods are emerging as a promising frontier of research, with several methods~\cite{song2025puppeteer,zhang2025unirig,Song_2025_CVPR} already achieving results comparable to the meticulous work of human artists. However, in stark contrast, methods for category-agnostic 3D animation generation—the final and crucial step in the asset creation workflow—remain in a nascent stage of research. This process continues to be a labor-intensive and time-consuming endeavor, predominantly reliant on manual craftsmanship, often requiring 3--7 days to produce just 10 seconds of animation. This disparity significantly impedes the overall efficiency of the 3D asset production pipeline.

Diverging from early methods that focused exclusively on category-specific motion generation, such as human motion generation~\cite{jiang2023motiongpt, tevet2022motionclip, zhang2024motiondiffuse, guo2022generating, song20213d, shen2025adhmr}, our work is category-agnostic to address the broader challenge of animating 3D models across arbitrary object categories. Within this paradigm, the research community's efforts have been predominantly channeled towards techniques based on mesh deformation animation, also known as 4D generation. Concretely, these mesh deformation methods create animations either by generating an entirely new 3D mesh for each frame~\cite{jiang2024animate3d,liang2024diffusion4d,uzolas2025motiondreamer,bahmani2024tc4d,chen2025v2m4} or predicting per-vertex displacements from the original mesh on a frame-by-frame basis~\cite{zhang2025gaussian,wu2025animateanymesh}. However, this approach is fundamentally constrained by inherent drawbacks. The prediction of mesh deformation is a computationally dense task, which often leads to slow inference speeds and pronounced animation jitter.

Facilitated by advancements in template-free rigging methods~\cite{song2025puppeteer,zhang2025unirig,Song_2025_CVPR} and the curation of new skeletal animation datasets~\cite{huang2025animax}, the research focus has recently pivoted towards the generation of skeletal animations. This paradigm has given rise to optimization-based methods. For instance, AKD~\cite{li2025articulated} employs Score Distillation Sampling (SDS) to refine joint kinematics, while Puppeteer~\cite{song2025puppeteer} leverages cues from tracking, optical flow, and depth as supervisory signals for the differentiable optimization of joint motion. Nevertheless, these approaches are encumbered by prohibitive computational demands, typically requiring tens of hours to synthesize a mere 100-frame animation. In a different vein, AnimaX~\cite{huang2025animax} adopts a strategy that requires extensive training to generate multi-view motion videos and subsequently employs triangulation and inverse kinematics to directly compute transformation matrices for individual joints. 

In this work, we present \textbf{AnimaSpark}, a highly efficient, feed-forward method for skeletal animation generation. Our approach is predicated on the assumption that for any given fundamental 3D motion, the corresponding joint transformations can be effectively modeled within a two-dimensional subspace. For instance, during the walking or running gaits of humans and quadrupeds, skeletal rotations are largely confined to the sagittal plane (as observed from a side view). Similarly, avian flight involves rotations predominantly in the coronal plane (as observed from a front view), while the swimming motions of fish or the propulsion of invertebrates occur primarily within the transverse plane (parallel to the ground). Building upon this foundational premise, our pipeline takes a rigged 3D model and a text prompt as input. The process commences by rendering a multi-layer image that encodes the model's joints, skeletal hierarchy, and mesh. Crucially, the rendering viewpoint should be approximately perpendicular to the plane in which the motion described in the text prompt occurs. The rendered image, conditioned on the text prompt, is then fed into a video generation model~\cite{wan2025,seedance} to synthesize a 2D motion sequence. Subsequently, we track the projected joints in the synthesized video. From these tracking results, we compute the per-frame 2D local rotation matrix for each joint, along with the translation matrix for the root joint. In the final stage, leveraging the initial rendering viewpoint and the extracted 2D motion data, we lift these 2D motion matrices back into the corresponding two-dimensional subspace within the 3D world to produce the final 3D animation.

Our primary contributions are twofold: (1) We introduce a novel, feed-forward pipeline for category-agnostic 3D animation that demonstrates superior performance in text-motion alignment, motion quality, and computational efficiency. (2) We propose a 2D-to-3D lifting formulation based on the key insight that the joint transformations underlying fundamental 3D motions can be effectively modeled within a two-dimensional subspace, enabling efficient inference without per-scene optimization.

\section{Related Work}
\label{sec:related}

\boldstartspace{Category-Agnostic 3D Animation Generation via Mesh Deformation.} This paradigm, often referred to as 4D generation, has made substantial progress in recent years. For instance, Animate3D~\cite{jiang2024animate3d} utilizes a multi-view video diffusion model (MV-VDM) to generate multi-view videos, which are then used in conjunction with 4D Score Distillation Sampling (4D-SDS) to reconstruct and refine the 3D animation. MotionDreamer~\cite{uzolas2025motiondreamer} decodes intermediate features from a pre-trained video diffusion model into discrete 3D objects, which are subsequently composed into the final animation. Similarly, V2M4~\cite{chen2025v2m4} adopts a frame-by-frame generation approach, first synthesizing an independent 3D object for each frame and then enhancing the animation's quality through extensive post-processing, such as mesh repositioning, appearance consistency optimization, and texture alignment. However, the strategy of generating a distinct 3D model for each frame results in prohibitively low inference efficiency. To address this issue, subsequent works like GVFDiffusion~\cite{zhang2025gaussian} and AnimateAnyMesh~\cite{wu2025animateanymesh} employ diffusion models to directly encode and predict inter-frame per-vertex displacements, improving inference speed by an order of magnitude. Nevertheless, per-vertex optimization remains a dense and challenging task, often introducing significant jitter and artifacts into the final animation. Furthermore, a fundamental limitation of this entire paradigm is that the animated object is often a new entity, distinct from the input model, which can lead to the loss of the intrinsic properties and appearance of the original input model.

\boldstartspace{Category-Agnostic 3D Animation Generation via Skeletal Animation.} Compared to mesh deformation, skeletal animation represents a more conventional and widely adopted paradigm in industrial production workflows. Facilitated by recent advancements in template-free rigging methods, the research community's focus has consequently pivoted towards the generation of category-agnostic 3D skeletal animations. This research direction has explored several distinct approaches. AnyTop~\cite{gat2025anytop}, for instance, leverages a diffusion model to learn motion priors that generalize across diverse skeletal structures. AKD~\cite{li2025articulated} employs video-based Score Distillation Sampling (SDS) to refine joint kinematics. Similarly, AnyMole~\cite{yun2025anymole} and Puppeteer~\cite{song2025puppeteer} both achieve end-to-end 3D animation generation by extracting skeletal animation from video through differentiable optimization. However, a significant drawback of these distillation and optimization-based strategies is the prohibitive computational burden they introduce, often requiring tens of hours to generate a mere 100-frame animation. Another noteworthy work in this domain is AnimaX~\cite{huang2025animax}. It trains a diffusion model to generate multi-view motion videos and subsequently employs triangulation and inverse kinematics to directly compute the transformation matrix for each joint. While this approach has the potential to achieve higher performance, its substantial data requirements present a considerable barrier to reproducibility and adoption.

\section{Method}
\label{sec:method}

\begin{figure*}
    \centering
\includegraphics[width=\textwidth]
{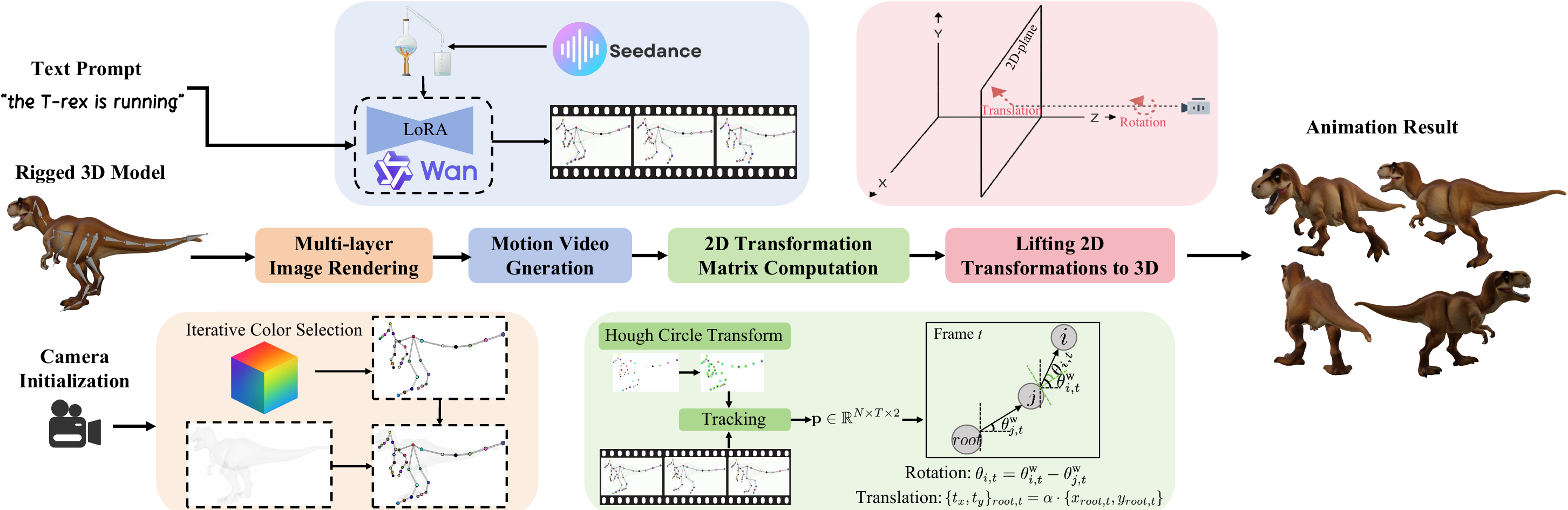}

    \caption{
    \textbf{Overview of our animation generation pipeline.} Our method consists of four main stages: 
        \textbf{(1) Multi-Layer Image Rendering:} Given a rigged 3D model, a text prompt, and a camera pose, we render a multi-layer image. 
        \textbf{(2) Motion Video Generation:} The image and prompt are fed into our fine-tuned model (Wan2.2~\cite{wan2025} adapted via distillation from Seedance-1.0-pro~\cite{seedance}) to synthesize a 2D motion video. 
        \textbf{(3) 2D Transformation Matrix Computation:} We track keypoints in the video and use their trajectories to compute the 2D transformation matrices. 
        \textbf{(4) Lifting 2D Transformations to 3D:} We lift the 2D matrices to 3D transformations and apply them to the skeleton to generate the final animation.
    } 
    \label{fig:overview}
  \end{figure*}

This section details our proposed method. We first formulate the problem of our skeletal animation generation task.
We then describe the four core components of our pipeline in sequence: Multi-Layer Image Rendering (Section~\ref{sec:method_rendering}), Motion Video Generation (Section~\ref{sec:method_video_generate}), 2D Transformation Matrix Computation (Section~\ref{sec:method_2D_matrix}), and finally, Lifting 2D Transformations to 3D (Section~\ref{sec:method_lifting}).
  
\vspace{-5pt}
\boldstartspace{Problem Formulation.} 
The goal of our method is to generate a skeletal animation for a given rigged 3D model, denoted as $\mathcal{M}$, conditioned on a textual prompt $\mathcal{T}$ and a set of camera pose parameters, denoted by $C$, used for intermediate rendering.
Formally, this task is defined as predicting a sequence of local transformation matrices, denoted as $M \in \mathbb{R}^{N \times T \times 4 \times 4}$, where $N$ represents the number of bones in the skeleton and $T$ is the total number of frames in the animation. For each bone $n \in \{1, ..., N\}$ at a specific frame $t \in \{1, ..., T\}$, its transformation is described by a $4 \times 4$ matrix $M_{n,t}$ that encapsulates the bone's rotation, translation, and scaling transformations:
\begin{equation}
    M_{n,t} = 
    \begin{bmatrix}
        r_{11}s_x & r_{12}s_y & r_{13}s_z & t_x \\
        r_{21}s_x & r_{22}s_y & r_{23}s_z & t_y \\
        r_{31}s_x & r_{32}s_y & r_{33}s_z & t_z \\
        0 & 0 & 0 & 1
    \end{bmatrix},
\end{equation}
where $r_{ij}$ represents an element of the $3 \times 3$ rotation matrix, while $\{s_x, s_y, s_z\}$ and $\{t_x, t_y, t_z\}$ are the components of the scaling and translation vectors, respectively.

Therefore, the overall process can be formulated as:
\begin{equation}
    M = F(\mathcal{M}, \mathcal{T}, C)
    \label{eq:high_level_formulation}
\end{equation}

\vspace{-5pt}
\subsection{Multi-Layer Image Rendering}
\label{sec:method_rendering}
For a given rigged 3D model $\mathcal{M}$, we first render it into a multi-layer image representation, denoted as $I$. 
This image, in conjunction with the textual prompt $\mathcal{T}$, serves as the conditional input for the subsequent video generation model, as detailed in Section~\ref{sec:method_video_generate}.
The rendering process begins by initializing a camera based on the predefined pose parameters $C$, wherein the camera's position is constrained to one of the cardinal axes (X, Y, or Z). 
This axis is selected such that the viewing direction is approximately perpendicular to the plane in which the motion described in the text prompt $\mathcal{T}$ primarily occurs.
We then render two distinct layers, $I_1$ and $I_2$.

The first layer, $I_1$, contains a semi-transparent rendering of the object's mesh. Its purpose is to provide the video generation model with visual cues regarding the object's category and overall structure.
The second layer, $I_2$, consists of an opaque rendering of the skeleton. The joints within this skeleton are rendered as distinct keypoints, which are to be tracked in the generated video. 
The skeletal hierarchy is visualized using grey lines to offer structural priors to the video generation model.
Finally, these two layers are composited into the final image $I$ by overlaying $I_2$ on top of $I_1$.

In the generated video, the rendered joints serve as the keypoints for tracking. However, they are susceptible to occlusion or may disappear due to hallucinatory artifacts from the video model, which can lead to tracking failure.
To mitigate this issue and enhance tracking robustness, we assign a unique and highly distinguishable color to each of the $N$ joints.
We devise an iterative color selection scheme to generate a set of $N$ maximally distant colors.
First, we generate a large pool of candidate colors, $\mathcal{C}$, by sampling uniformly from the RGB color space $[0,1]^3$, and filter it to remove achromatic colors (i.e., those close to black, white, and the grayscale axis).
The set of final joint colors, $\mathcal{S}$, is then built iteratively. After initializing $\mathcal{S}$ with one random color, the next color $\mathbf{c}^*$ is chosen from the remaining candidates by finding the one that maximizes the minimum Euclidean distance to all previously selected colors:
\begin{equation}
    \mathbf{c}^* = \arg\max_{\mathbf{c} \in \mathcal{C} \setminus \mathcal{S}} \left( \min_{\mathbf{c}_j \in \mathcal{S}} \|\mathbf{c} - \mathbf{c}_j\|_2 \right)
    \label{eq:color_selection}
\end{equation}

This process is repeated until $N$ distinct colors are selected.

\vspace{-5pt}
\subsection{Motion Video Generation}
\label{sec:method_video_generate}
We generate a video sequence, denoted as $V$, conditioned on the text prompt $\mathcal{T}$ and the multi-layer image $I$.
During our initial experiments, we observed a high frequency of failures with standard pre-trained video generation models. 
This issue was prevalent in both open-source models, such as Wan2.2~\cite{wan2025}, and more powerful, closed-source models like Seedance-1.0-pro~\cite{seedance}. 
Common failure modes included: (1) blurred or disappearing joints; (2) the hallucination of extraneous joints; (3) asynchronous motion between the skeleton and the mesh; and (4) poor adherence to the textual motion prompt. 
For Seedance-1.0-pro, such failures occurred in approximately 60\% of cases, while the failure rate for Wan2.2 exceeded 90\%. 
We attribute this phenomenon to a domain gap: the style of our multi-layer image $I$ is likely too unfamiliar for these models to comprehend the complex semantics and structural relationships between the mesh and skeleton.

To address this challenge, we adopted a targeted fine-tuning strategy. We first conducted extensive experiments on the more capable Seedance-1.0-pro model to curate a high-quality dataset of successful generation results. 
Ultimately, we collected videos spanning 61 distinct motions across 20 object categories. 
We then used this curated dataset to fine-tune the Wan2.2 model via Low-Rank Adaptation (LoRA). 
After this fine-tuning process, the adapted Wan2.2 model demonstrated a significant improvement.


\vspace{-5pt}
\subsection{2D Transformation Matrix Computation}
\label{sec:method_2D_matrix}

\vspace{-5pt}
\boldstartspace{Keypoint Tracking.}
We perform keypoint tracking on the generated motion video $V$ to acquire the positional information of each joint in every frame.
This process begins with detecting the initial positions of all joints in the multi-layer image $I$. 
For this, we employ the Hough Circle Transform, an algorithm designed to detect circles in an image. 
This yields a set of bounding boxes $\mathcal{B} = \{B_1, \dots, B_N\}$ for all $N$ joints.
Since the image $I$ also serves as the first frame of the video $V$, we utilize Cotracker3~\cite{karaev2024cotracker3} to track the center points of $\mathcal{B}$ throughout the video $V$, which yields the trajectory for each joint $\mathbf{p} \in \mathbb{R}^{N \times T \times 2}$, where each element $\mathbf{p}_{n,t}$ represents the 2D coordinates of the $n$-th joint at frame $t$.

\vspace{-5pt}
\boldstartspace{Transformation Matrix on the Plane.}
In our Multi-Layer Image Rendering stage, the camera pose parameters $C$ are constrained to a viewpoint along one of the cardinal axes (X, Y, or Z). 
Consequently, the video plane corresponds to the plane formed by the remaining two axes.
This setup allows us to employ Forward Kinematics to directly derive the transformation components for each joint on these two axes from the tracking results $\mathbf{p}$.
For instance, if the camera is positioned along the Z-axis and the motion occurs primarily on the XY-plane, the 2D local transformation matrix on the XY-plane can be expressed as:
\begin{equation}
    M_{2D, n,t} = 
    \begin{bmatrix}
        r_{11}s_x & r_{12}s_y &  t_x \\
        r_{21}s_x & r_{22}s_y &  t_y \\
        0 & 0 & 1  \\
    \end{bmatrix}
    \label{eq:2d_transform_matrix}
\end{equation}

Without loss of generality, our subsequent discussion will assume this case where the camera is positioned along the Z-axis and the motion occurs primarily on the XY-plane.
Building on this, we fix the scaling factor for all joints to 1 and then employ Forward Kinematics to compute the 2D rotation for each joint and the 2D translation for the root joint.

\vspace{-5pt}
\boldstartspace{Rotation.} At each frame $t$, for a given joint $i$ and its parent $j$, we compute the global orientation angle of the bone, $\theta_{i,t}^{\text{w}}$, from their tracked 2D positions, $\mathbf{p}_{i,t}=(x_{i,t}, y_{i,t})$ and $\mathbf{p}_{j,t}=(x_{j,t}, y_{j,t})$, using the two-argument arctangent function:
\begin{equation}
    \theta_{i,t}^{\text{w}} = \text{atan2}(y_{j,t} - y_{i,t}, x_{j,t} - x_{i,t})
    \label{eq:angle_computation}
\end{equation}

Subsequently, by substituting the rotation components $r_{11}, r_{12}, r_{21}, r_{22}$ in the transformation matrix with their trigonometric definitions ($\cos\theta, -\sin\theta, \sin\theta,$ and $\cos\theta$, respectively), the world transformation matrix for joint $i$ at frame $t$, $M_{2D, i,t}^{\text{w}}$, can be expressed as:
\begin{equation}
M_{2D, i,t}^{\text{w}} = 
\begin{bmatrix}
    R(\theta_{i,t}^{\text{w}}) & \mathbf{0} \\
    \mathbf{0}^T & 1
\end{bmatrix},
\qquad
R(\theta_{i,t}^{\text{w}}) = 
\begin{bmatrix}
    \cos \theta_{i,t}^{\text{w}} & -\sin \theta_{i,t}^{\text{w}} \\
    \sin \theta_{i,t}^{\text{w}} & \cos \theta_{i,t}^{\text{w}}
\end{bmatrix}
\label{eq:maxtri_world}
\end{equation}

According to the principles of skeletal animation, the world transformation matrix for joint $i$ ($M_{2D, i,t}^{\text{w}}$), the world transformation matrix for its parent $j$ ($M_{2D, j,t}^{\text{w}}$), and the local transformation matrix for joint $i$ ($M_{2D, i,t}$) adhere to the following relationship: 
\begin{equation}
    M_{2D, i,t}^{\text{w}} = M_{2D, j,t}^{\text{w}} \cdot M_{2D, i,t}
\label{eq:maxtri_relation}
\end{equation}

Substituting Eq.~\ref{eq:maxtri_world} into Eq.~\ref{eq:maxtri_relation} yields:
\begin{equation}
    R(\theta_{i,t}^{\text{w}}) = R(\theta_{j,t}^{\text{w}}) R(\theta_{i,t}) = R(\theta_{j,t}^{\text{w}} + \theta_{i,t}),
    \label{eq:rotation_relation}
\end{equation}
which implies:
\begin{equation}
    \theta_{i,t} = \theta_{i,t}^{\text{w}} - \theta_{j,t}^{\text{w}}
    \label{eq:angle_relation}
\end{equation}

Consequently, based on this derived relationship, the local rotation $R(\theta_{n,t})$, an essential component of the final 2D local transformation matrix $M_{2D, n,t}$, can be efficiently calculated via a two-argument arctangent function on the tracking results $\mathbf{p}$ and a simple subtraction.

\vspace{-5pt}
\boldstartspace{Translation.} For the root joint, its translation vector at each frame $t$, denoted as $\{t_x, t_y\}_{root,t}$, is directly given by its tracked 2D position, $\mathbf{p}_{root,t}$:
\begin{equation}
    \{t_x, t_y\}_{root,t} = \alpha \cdot \{x_{root,t}, y_{root,t}\},
\end{equation}
where $\alpha$ is the scaling factor.

Based on the derivations above, we recursively compute the 2D local transformation matrix $M_{2D, n,t}$ for every joint, resulting in a final tensor $M_{2D} \in \mathbb{R}^{N \times T \times 3 \times 3}$. 
Subsequently, Gaussian smoothing is applied along the temporal dimension ($T$) of this tensor to ensure the smoothness of the final animation.

\vspace{-5pt}
\subsection{Lifting 2D Transformations to 3D}
\label{sec:method_lifting}
Since the rendering viewpoint is deliberately selected to be approximately perpendicular to the primary plane of motion, discarding the transformation components along the third axis does not significantly degrade the final animation quality. 
Specifically, for each $3 \times 3$ 2D local transformation matrix $M_{2D,n,t}$, we lift it to a 3D local transformation matrix $M_{n,t}$ by embedding it into a $4 \times 4$  identity matrix:
\begin{equation}
    M_{2D, n,t} = 
    \begin{bmatrix}
        r_{11}s_x & r_{12}s_y &  t_x \\
        r_{21}s_x & r_{22}s_y &  t_y \\
        0 & 0 & 1 \\
    \end{bmatrix}
    \quad \Rightarrow \quad
    M_{n,t} = 
    \begin{bmatrix}
        r_{11}s_x & r_{12}s_y & 0 & t_x \\
        r_{21}s_x & r_{22}s_y & 0 & t_y \\
        0 & 0 & 1 & 0 \\
        0 & 0 & 0 & 1
    \end{bmatrix}
\end{equation}

Ultimately, the per-joint, per-frame transformation matrices $M_{n,t}$ are aggregated into a final tensor $M \in \mathbb{R}^{N \times T \times 4 \times 4}$, which is then applied to the skeleton of the rigged model $\mathcal{M}$ to drive the animation.

\vspace{-5pt}
\section{Experiments}
\label{sec:exp}
\vspace{-5pt}
\subsection{Experimental Setup}
\vspace{-5pt}
\boldstartspace{Baselines and Metrics.} 
We compare our method against one skeleton-based baseline, Puppeteer~\cite{song2025puppeteer}, and four mesh deformation-based baselines: GVFDiffusion~\cite{zhang2025gaussian}, V2M4~\cite{chen2025v2m4}, MotionDreamer~\cite{uzolas2025motiondreamer}, and AnimateAnyMesh~\cite{wu2025animateanymesh}.
For both quantitative and qualitative evaluation, we selected a diverse set of six 3D models, either generated by Hunyuan3D 2.5~\cite{lai2025hunyuan3d25highfidelity3d} and Tripo~\cite{tripo3d}, or collected from the internet. These models span various categories, including humans, quadrupeds, bipeds, birds, and plants. Each model was then paired with a corresponding text prompt to generate an animation.
Our quantitative evaluation follows the methodology of Huang et al.~\cite{huang2025animax}. The process involves rendering the generated 3D animations into videos and subsequently evaluating their quality using the VBench benchmark~\cite{huang2023vbench}.

\vspace{-5pt}
\boldstartspace{Implementation Details.} 
A common paradigm shared by our method and selected baselines is a two-stage process: first generating an intermediate video, and then extracting the 3D animation from it.
However, different methods require input videos with distinct visual styles. 
For instance, our method utilizes the skeleton and semi-transparent mesh style shown in Figure~\ref{fig:overview}, whereas Puppeteer requires photorealistic appearance. 
This makes it infeasible to use a single, identical video to generate animations across all methods.
To ensure a fair and controlled evaluation, we implemented a rigorous selection protocol. For each target motion, we iteratively generated multiple videos for every required style. This process was continued until we could curate a set of videos that, while stylistically diverse, exhibited motions that were qualitatively nearly identical.

\vspace{-5pt}
\subsection{Qualitative Comparisons}
\begin{figure*}[!htb]
    \centering
\includegraphics[width=\textwidth]
{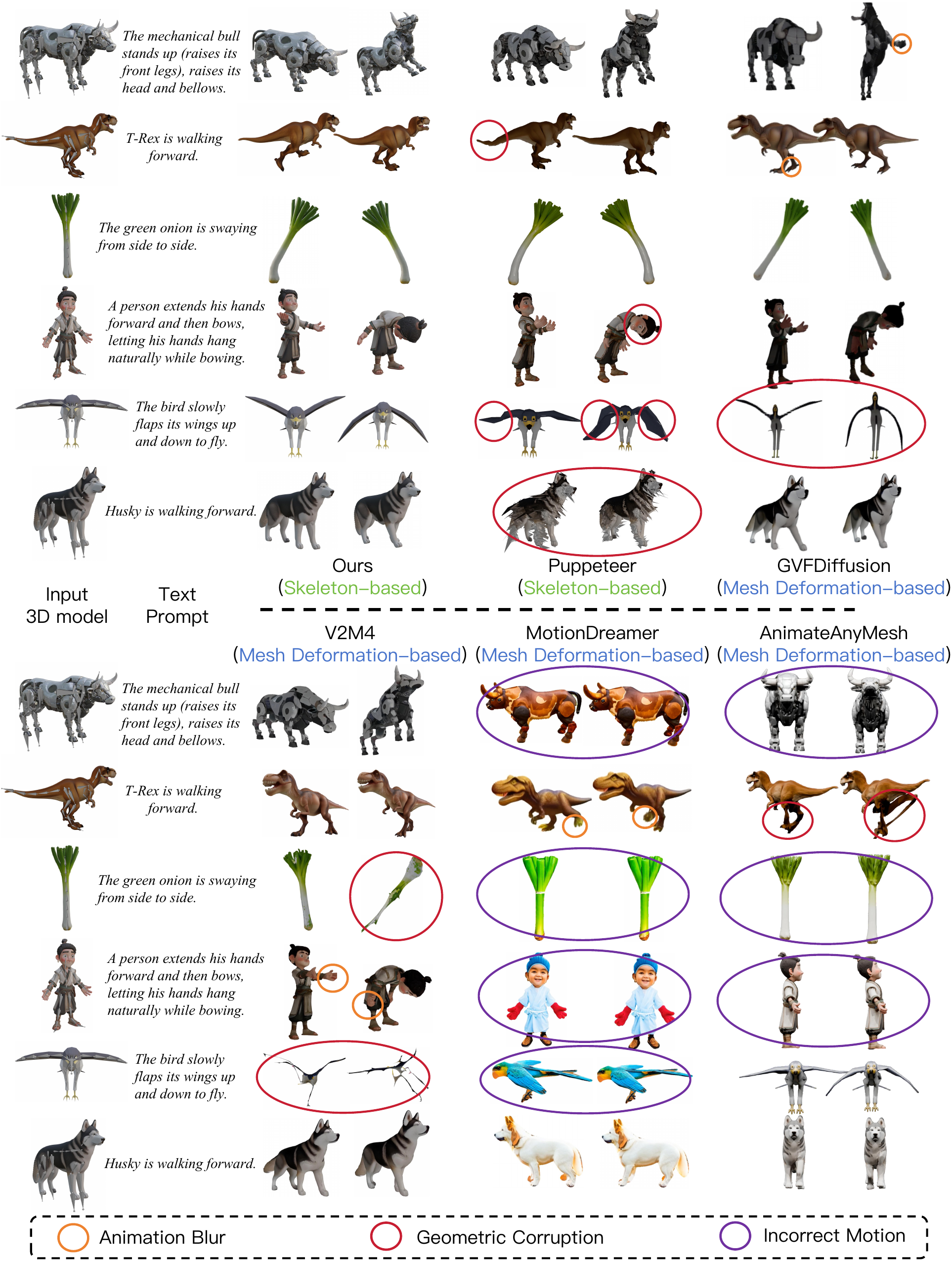}

    \caption{Qualitative comparison with state-of-the-art skeleton-based and mesh deformation-based methods. In the generated results, we annotate failure modes such as animation blur, geometric corruption, and incorrect motion with colored circles. 
    }
    \label{fig:fig3}
 \vspace{-10pt}
  \end{figure*}
We conduct a qualitative comparison with several state-of-the-art methods, including Puppeteer~\cite{song2025puppeteer}, GVFDiffusion~\cite{zhang2025gaussian}, V2M4~\cite{chen2025v2m4}, MotionDreamer~\cite{uzolas2025motiondreamer}, and AnimateAnyMesh~\cite{wu2025animateanymesh}. As shown in Figure~\ref{fig:fig3}, despite multiple generation attempts and selecting the best outcomes, the mesh deformation-based methods exhibit significant limitations. 
The common strategy of generating a new 3D mesh frame-by-frame or performing dense per-vertex optimization inevitably leads to animation blur and jitter.
At the same time, dense per-vertex optimization is particularly unreliable, often resulting in severe geometric corruption, as evidenced by the outputs of GVFDiffusion, V2M4, and AnimateAnyMesh.
Another limitation arises in methods such as MotionDreamer, which reconstruct the 3D model on a per-frame basis from the intermediate features of a video model. The potential information loss in this decoding process can cause nearly static or otherwise incorrect motion. The issue of incorrect motion in AnimateAnyMesh is compounded by its technical limitation of generating only 16-frame animations, a duration insufficient for depicting most complex movements.

While Puppeteer's use of skeletal animation effectively mitigates blur and some forms of incorrect motion, the model is still prone to geometric corruption. This issue arises because Puppeteer extracts skeletal motion via differentiable optimization on a target video, using a loss function that combines tracking, optical flow, and depth signals. These signals, however, can be unreliable. For instance, the skeleton of the input 3D model in Figure~\ref{fig:fig3} extends beyond the mesh boundaries. Consequently, when the object motion video is rendered, certain bones are occluded or lost from view. Tracking these bones subsequently fails, leading to a loss of local motion for those corresponding parts. Unlike these baseline methods that often fail to produce correct animations despite repeated attempts, our approach reliably generates a usable result within just a few trials. This efficiency is critical, as baseline methods require hours per animation, making retrials prohibitively expensive. In stark contrast, our method takes only 3 min per inference, dramatically lowering the cost and effort of regeneration.

\begin{figure*}[!htb]
    \centering
\includegraphics[width=\textwidth]
{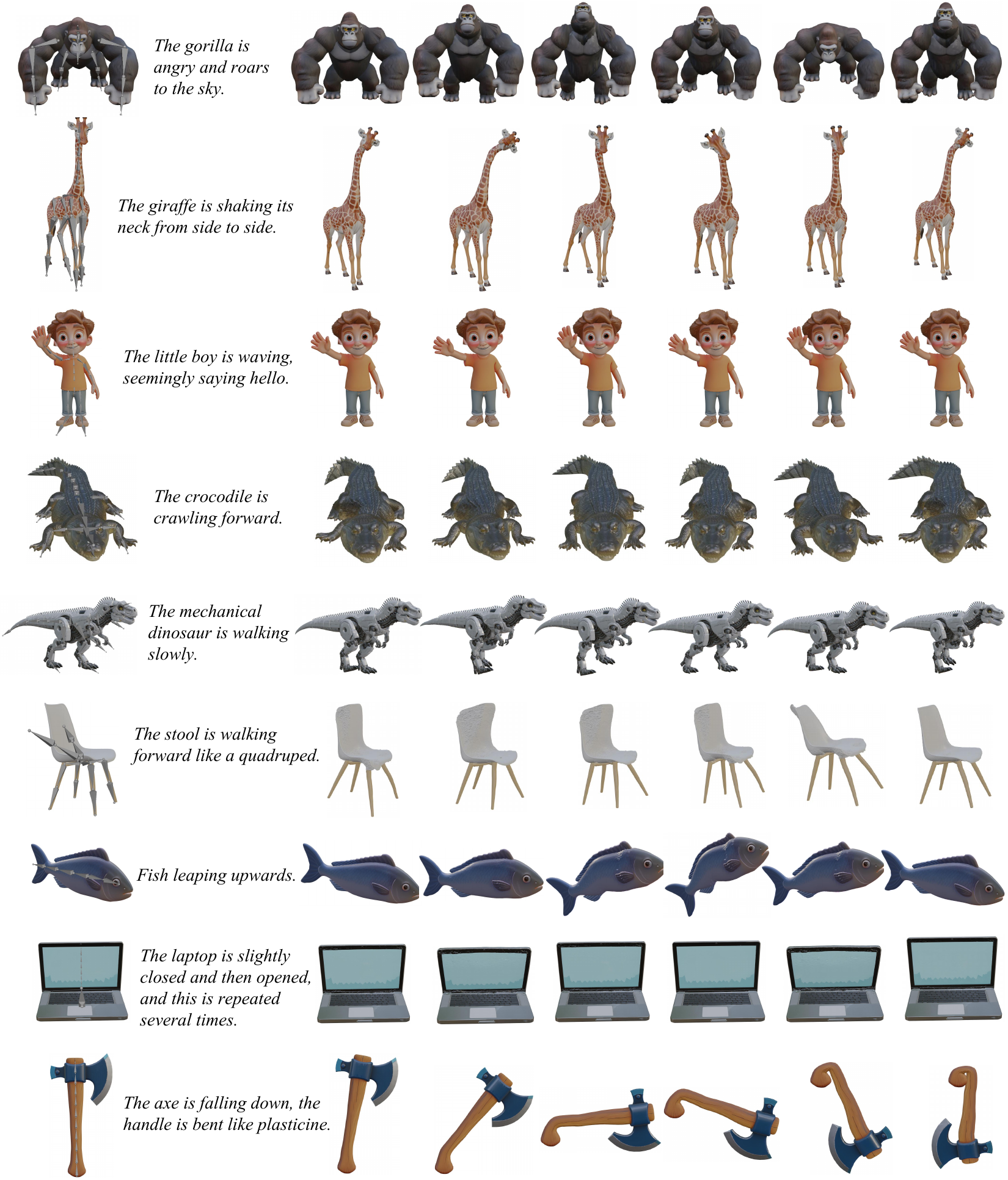}
    \caption{Additional visualizations from our method.}
    \label{fig:fig4}
     \vspace{-10pt}
  \end{figure*}

For more visualizations, please refer to Figure~\ref{fig:fig4}.

\vspace{-5pt}
\subsection{Quantitative Comparisons}

We render our generated animations into videos and employ VBench, an automated tool for video quality assessment, to measure three key metrics: subject consistency, motion smoothness, and appearance quality. These metrics serve as effective proxies for the underlying quality of the animation. The quantitative results, presented in Table~\ref{tab:1}, show that our method outperforms the competing baselines across all evaluated metrics.

{
\renewcommand{\arraystretch}{1.2}
\begin{table*}[h]
  \centering
  \vspace{-5pt}
  \caption{Quantitative results of subject consistency, motion smoothness and appearance quality using the VBench~\cite{huang2023vbench} benchmark.}
  \setlength{\tabcolsep}{11pt}
  \begin{tabular}{cccc}
    \toprule
    \multirow{2}{*}{Method} &
    \multirow{2}{*}{\shortstack{Subject \\ Consistency$\uparrow$}} & 
    \multirow{2}{*}{\shortstack{Motion \\ Smoothness$\uparrow$}} & 
    \multirow{2}{*}{\shortstack{Appearance \\ Quality$\uparrow$}} \\
    \\
    \midrule 
    AnimateAnyMesh~\cite{wu2025animateanymesh}  & 0.8387 & 0.9665 & 0.4595 \\
    MotionDreamer~\cite{uzolas2025motiondreamer}  & 0.8798 & 0.9731 & 0.4828 \\
    V2M4~\cite{chen2025v2m4}  & 0.8274 & 0.9715 & 0.4718 \\
    GVFDiffusion~\cite{zhang2025gaussian}  & 0.8811  & 0.9878 & 0.4873 \\
    Puppeteer~\cite{song2025puppeteer}  & 0.9297 & 0.9806& 0.5196  \\
    \midrule
    Ours   &  \textbf{0.9472} & \textbf{0.9954} & \textbf{0.5243}   \\
    \bottomrule
  \end{tabular}
  \vspace{-5pt}
    \label{tab:1}
\end{table*}
}

As analyzed in Table~\ref{tab:2}, we compare the inference efficiency and additional input requirements of our method against several baselines. While AnimateAnyMesh offers fast inference and does not require an initial rendering angle or an intermediate video, its utility is severely limited by a 16-frame animation length, which is too short for most practical applications. MotionDreamer, V2M4, and GVFDiffusion also eliminate the need for an initial angle but are significantly slower than our approach. For instance, V2M4 requires 5 hours for a single generation compared to our 3 minutes.
The most direct comparison is with Puppeteer, which shares our requirement for an initial angle and a video generation step. Critically, our method is far more efficient, requiring only 0.5\% of Puppeteer's inference time. In a practical workflow, even if we run our method multiple times to find the best result and assume Puppeteer succeeds on its first try, our total time commitment is merely 1--5\% of theirs. This dramatic speed-up significantly improves the viability of our method for iterative and real-world use cases.

{
\renewcommand{\arraystretch}{1.2}
\begin{table*}[h]
  \centering
  \vspace{-5pt}
  \caption{Inference efficiency and additional requirements. The inference times are measured by averaging the results of multiple runs on the same GPU.}
    \setlength{\tabcolsep}{9pt}
  \begin{tabular}{ccccc}
    \toprule
    Method  & Inference Time & Length Limit  & Video Step & Angle Input  \\
    \midrule 
    AnimateAnyMesh~\cite{wu2025animateanymesh}  & 3 min & \ding{51}  & \ding{55} & \ding{55} \\
    MotionDreamer~\cite{uzolas2025motiondreamer}  & 1 h  & \ding{55}  & \ding{51} & \ding{55} \\
    V2M4~\cite{chen2025v2m4}  & 5 h  & \ding{55}  & \ding{51} & \ding{55} \\
    GVFDiffusion~\cite{zhang2025gaussian}  & 10 min & \ding{55}  & \ding{51} & \ding{55} \\
    Puppeteer~\cite{song2025puppeteer}  & 10 h  & \ding{55}  & \ding{51} & \ding{51} \\
    \midrule
    Ours   &  3 min & \ding{55}  & \ding{51} & \ding{51}  \\
    \bottomrule
  \end{tabular}
  \vspace{-5pt}
\label{tab:2}
\end{table*}
}

\vspace{-5pt}
\subsection{User Study}

We conducted a blind user preference study with 28 participants to compare our method against competing approaches. 
For each case presented in Figure~\ref{fig:fig3}, participants were shown 6 animations generated from the same input 3D model and text prompt. 
The source of each animation was hidden, and their presentation order was randomized to ensure a fair comparison. 
For each set, participants were asked to select the best animation for each of the following criteria:
\vspace{-5pt}
\begin{enumerate}
    \item Text-Animation Alignment: Which animation best matches the input text description?
    \item Motion Naturalness: Which animation's object motion appears most natural and physically plausible?
    \item Geometric Consistency: Which animation best preserves the 3D shape with the fewest geometric artifacts (such as jitter, deformation, or unnatural creases)?
\end{enumerate}
  \vspace{-5pt}
As summarized in Table~\ref{tab:user_study}, the results show a consistent preference for our method across all three evaluation metrics.

{
\renewcommand{\arraystretch}{1.2}
\begin{table*}[h]
  \centering
  \vspace{-5pt}
  \caption{User study results. The results from 28 participants demonstrate our method's superiority over competing approaches, based on preference ratings for text-animation alignment, motion naturalness, and 3D geometric consistency.}
    \setlength{\tabcolsep}{9pt}
  \begin{tabular}{cccc}
    \toprule
    \multirow{2}{*}{Method} &
    \multirow{2}{*}{\shortstack{Text-Animation \\ Alignment$\uparrow$}} & 
    \multirow{2}{*}{\shortstack{Motion \\ Naturalness$\uparrow$}} & 
    \multirow{2}{*}{\shortstack{3D Geometric \\ Consistency$\uparrow$}} \\
    \\
    \midrule 
    AnimateAnyMesh~\cite{wu2025animateanymesh}  & 1.15\% & 0.00\% & 1.20\% \\
    MotionDreamer~\cite{uzolas2025motiondreamer}  & 1.72\%  & 4.76\% & 0.60\% \\
    V2M4~\cite{chen2025v2m4}  & 10.92\%  & 10.71\%  & 5.39\%\\
    GVFDiffusion~\cite{zhang2025gaussian}  & 7.47\%  & 7.14\%  & 14.97\%  \\
    Puppeteer~\cite{song2025puppeteer}  & 16.10\% & 16.07\%  & 11.98\% \\
    \midrule
    Ours   &  \textbf{62.64\%} & \textbf{61.31\%} & \textbf{65.87\%}   \\
    \bottomrule
  \end{tabular}
  \vspace{-5pt}
\label{tab:user_study}
\end{table*}
}

\section{Conclusion}
\vspace{-5pt}
In this work, we introduce a novel feed-forward pipeline for category-agnostic 3D animation generation. Our pipeline comprises four distinct stages: Multi-Layer Image Rendering, Motion Video Generation, 2D Transformation Matrix Computation, and Lifting 2D Transformations to 3D. This design enables the generation of skeletal animations for diverse basic motions across a diverse range of categories, all while achieving fast inference. When compared to state-of-the-art methods, our approach exhibits superior performance in terms of both animation quality and inference speed, presenting a compelling new direction for efficient and versatile 3D animation generation.

\section*{Limitations and Future Work} 
The primary limitation of our method stems from its core assumption that joint transformations for basic motions can be modeled in a two-dimensional subspace. This restricts its application to relatively simple actions. For complex motions, such as an intricate dance, motion along all three spatial dimensions may be crucial for capturing the full performance.
To address this, we plan to extend our method to capture full 3D motion by incorporating a differentiable optimization module, similar to that used in Puppeteer, specifically to model motion in the third dimension (i.e., along the axis perpendicular to the camera plane). Unlike Puppeteer, which applies this costly optimization to all dimensions, our proposed hybrid approach decouples the problem: we use efficient 2D forward kinematics for the main motion plane and reserve 1D differentiable optimization only for the depth axis. This strategy is designed to preserve 3D motion expressiveness while avoiding the prohibitively long inference times of a purely optimization-based approach.

\bibliography{Reference}

\end{document}